\documentclass[conference,a4paper]{IEEEtran}
\IEEEoverridecommandlockouts

\usepackage{cite}
\usepackage{amsmath,amssymb,amsfonts}
\usepackage{algorithmic}
\usepackage{graphicx}
\usepackage{textcomp}
\usepackage{xcolor}
\def\BibTeX{{\rm B\kern-.05em{\sc i\kern-.025em b}\kern-.08em
    T\kern-.1667em\lower.7ex\hbox{E}\kern-.125emX}}
\usepackage[T1]{fontenc}
\usepackage{xcolor}
\usepackage[normalem]{ulem}
\usepackage{graphicx}
 \usepackage{amsmath} 
 \usepackage{subcaption}
 \usepackage{multirow}
 \usepackage{url}
 \usepackage{soul}
 \newcommand{\add}[1]{\textcolor{black}{#1}}
 \usepackage{flushend}
 \def\error{\text{*}}
  \def\e{\epsilon}

\def\myexchange#1#2{#2}
\def\LLM{ChatGPT}
\def\GenAI{ChatGPT}
\def\mfai{my friend and I}
\def\sotaasr{{\tt \small whisper}}
\def\asrclm{{\sc asr-clm}}
\def\gs{{\sc g-score}}
\def\wer{{\sc wer}}

\def\C{P_s}
\def\X{S(t)}
\def\P{p}
\def\clm{{\sc clm}}
\def\ASR{{\sc asr}}

\def\Della{ \#1}
 \def\Upasana{ \#2}
 \def\Samina{ \#3}
 \def\Ashwini{ \#4}
 \def\sSunilzerozero{ \#5}
 \def\sSunilzeroone{ \#6}
 \def\sSunilzerotwo{ \#7}
 \def\onesevenonefivefivesevenzeroone{ \#8}
  
 \def\fiveonenineseversixsixzeroone{ \#10}
\def\fiveonetwosevensixsixzerotwo{ \#11}
\def\fiveonenineseversixsixzerothree{ \#12} 
\def\fiveonenineseversixsixzerofour{ \#13} 
\def\eightfiveeightseventhreezerozeroone{ \#14}
\def\prajwalzeroone{ \#15} 
\def\rajulzeroone{ \#16}
\def\Thambidurai{ \#17}

\def\sla{SLA}
\def\wla{WLA}
\def\ltwo{L2}

\begin{document}
%
 \title{Spoken Grammar Assessment Using LLM}
%
%
\author{Sunil Kumar Kopparapu
\and Chitralekha Bhat
\and Ashish Panda
}
%
%


\author{\IEEEauthorblockN{Sunil Kumar Kopparapu}
\IEEEauthorblockA{\textit{TCS Reserach} \\
\textit{Tata Consultancy Services Limited}\\
Mumbai, India \\
sunilkumar.kopparapu@tcs.com}
\and
\IEEEauthorblockN{Chitralekha Bhat}
\IEEEauthorblockA{\textit{TCS Reserach} \\
\textit{Tata Consultancy Services Limited}\\
Mumbai, India \\
chitralekha.bhat@gmail.com}
\and
\IEEEauthorblockN{Ashish Panda}
\IEEEauthorblockA{\textit{TCS Reserach} \\
\textit{Tata Consultancy Services Limited}\\
Mumbai, India \\
ashish.panda@tcs.com}
}

\maketitle              
\begin{abstract}
Spoken language assessment (\sla) systems restrict themselves to evaluating the pronunciation and oral fluency of a speaker by analysing the read and spontaneous spoken utterances respectively. 
The assessment of language grammar or vocabulary is relegated to written language assessment (\wla) systems. 
Most \wla\ systems present a set of sentences from a curated finite-size database of sentences thereby making it possible to anticipate the test questions and train oneself.  
In this paper, we propose a novel end-to-end \sla\ system to assess
 language grammar from spoken utterances thus making \wla\ systems redundant; additionally, we make the assessment largely unteachable by employing a large language model (LLM) to bring in variations in the test. We further demonstrate that a hybrid automatic speech recognition (\ASR) with a custom-built language model outperforms the state-of-the-art \ASR\  engine for spoken grammar assessment. 
\end{abstract}


%
%
%
\section{Introduction}
\label{sec:introduction}
 
 The demand for second language (\ltwo) learners to study foreign languages, especially English, leads to the imminent need for the development of language proficiency assessment systems or tools \cite{jin2023developing,franco2010eduspeak}. While several English language assessment tools exist, the
 assessments are often lengthy because they have separate assessment modules to assess different aspects of language proficiency. The spoken language proficiency assessment is often restricted to assessing the speech articulation of the speaker in terms of pronunciation \cite{sheoran2023pronunciation,DBLP:journals/taslp/PeiFLX24,lin2023exploiting} and speech delivery in terms of oral fluency \cite{preciado2018speaker,dubagunta2022towards}, which includes speaking rate \cite{imran2013speakrite,kopparapu2015non}, recognition of pauses, filler words, and analysis of intonation \cite{arias2010automatic} etc. The other important aspects of language like grammar or vocabulary are assessed separately through a written language proficiency assessment. 
 Spoken language assessment (\sla) and written language assessment (\wla) complement each other, providing a comprehensive evaluation of overall language proficiency.
 Separate \sla\ and \wla\ assessments not only extend testing time but may also encourage learners to neglect grammar. In practical settings like call centers and virtual interviews, spoken language communication is important. This highlights the need for a comprehensive \sla\ system that assesses all aspects of language proficiency. The primary obstacle to integrating grammar assessment into current \sla\ systems is the limited availability or accuracy of speech analysis tools. Accurate grammar assessment requires precise identification of spoken words by ASR engines, which can be challenging due to the limitations of ASR, especially with spontaneous speech. As a result, grammar assessment is often delegated to \wla\ systems. For further insights on the differences between spoken and written language, refer to
 For details on the difference between spoken and written language text see \cite{9806803}.  
 
 \add{The study \cite{10446782} compared a cascaded system with separate modules for \ASR\, disfluency removal, and grammar error correction, to an end-to-end system and demonstrated that the performance of the latter system was comparable to that of the former. With current advancements in \ASR\ technology, often it can be believed that these systems could capture spoken grammatical errors in the decoded text. However, these systems have an inherent bias from the language model (LM) towards the grammatically correct text. The study \cite{8683080} found that a deep learning-based grammatical error detection (GED) system, fine-tuned on free speech data, improved performance on non-native spoken English. However, challenges in \ASR\ and disfluency detection limited accurate feedback. The work \cite{Lu2019ImpactOA} evaluated the impact of \ASR\ errors on GED using a deep learning-based system originally trained on written text. \ASR\ confidence scores were integrated into the GED system to address the grammatical errors stemming from incorrect transcriptions rather than learner mistakes.  In \cite{venkatathirumalakumar23_slate}, the authors evaluated \ASR\-based methods for spoken GED, finding that a score-based classification outperforms the cascaded approach. They also found that LM and N-best hypotheses had minimal impact on decoding-based likelihood classification. The above two studies highlight the issue with the current spoken GED systems that use SOTA \ASR\ and the need for a system using custom-built LMs.}

 In this paper, we introduce an end-to-end \sla\ system 
 to enable \add{GED} or assessment of language grammar from spoken speech.
 Further, the use of a large language model (LLM) makes the \sla\ system scalable and practical because 
 no two assessment instances are the same; ensuring that the student cannot be coached for the assessment. The main contribution 
 of the paper 
 is (a) designing a \sla\ system 
 that can robustly evaluate all aspects of language proficiency, without employing additional \wla\ tools, thereby significantly reducing the time taken to take the test, (b) proposing a mechanism to incorporate language grammar assessment by exploiting the superior performance of available speech analysis tools on read speech, (c) automatic grammar assessment using a custom-built LM on top of a readily available hybrid \ASR\  system, (d) proposing a grammar scoring module that is robust to errors in \ASR, and (e) 
 employing LLM to to bring in variations in the test to make the \sla\ system largely unteachable thus making it scalable and practical.  
 The rest of the paper is organized as follows, we describe the spoken language grammar assessment system in detail in Section \ref{sec:overview}. We conduct experiments in Section \ref{sec:experimental-results} to show the process of automatic generation of paragraphs that can be used in grammar evaluation and show the need for a custom-built LM for speech transcription and we conclude in Section \ref{sec:conclusions}.

\section{Spoken Language Grammar Assessment}
\label{sec:overview}

\begin{figure}
   
    \begin{subfigure}{0.5\textwidth}
     \centering
        \includegraphics[width=0.9\linewidth]{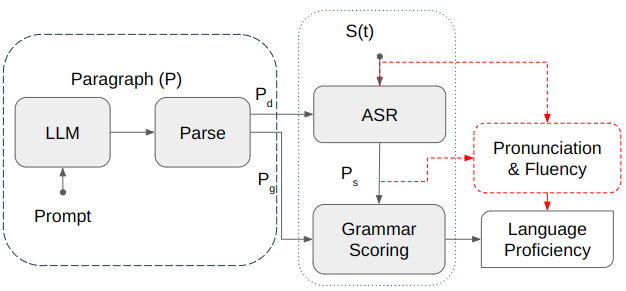}
        \caption{Block Diagram}
        \label{fig:e2e-system}
    \end{subfigure}
    \begin{subfigure}{0.5\textwidth}
    \centering
    \includegraphics[width=0.75\linewidth]{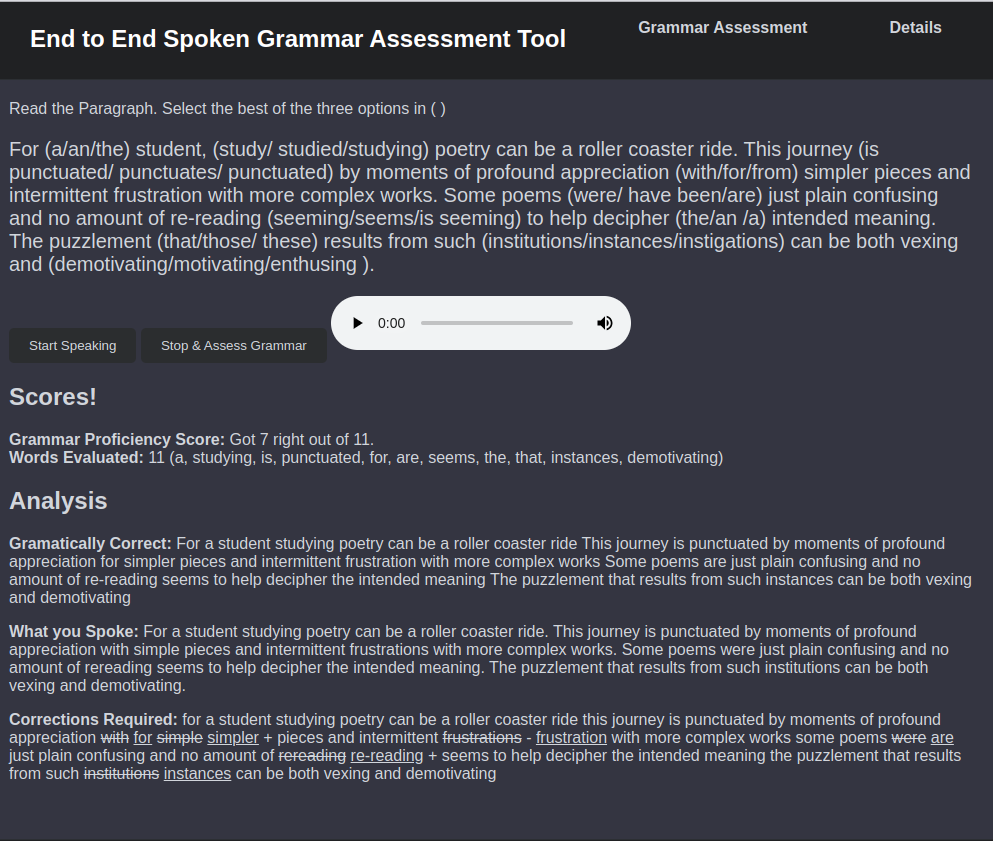}
    \caption{Functional System (Web Application).}
    \label{fig:web_interface}
    \end{subfigure}
     \caption{End to End System for \sla. We only look at the grammar of spoken language.}
\end{figure}

The block diagram of the end-to-end \sla\ system is shown in Fig. \ref{fig:e2e-system}. It has two parts, the first part, allows for the generation of a paragraph $P$ (example Fig. \ref{fig:gen-paragraph}(a)) by prompting an LLM, and the second part takes 
the audio $S(t)$, spoken by the candidate, corresponding to $P_d$ (example, \ref{fig:gen-paragraph}(b))  and assesses for language grammar using $P_g$ (Fig. \ref{fig:gen-paragraph}(c)). 
Unlike traditional \sla\ systems which take an audio input $S(t)$  and use the output $P_s$ of a standard \ASR\  to automatically compute the pronunciation or oral fluency \cite{DBLP:conf/eusipco/PandaAK23,fontan:fluency} only,  in this paper, 
we enable grammar assessment on spoken speech. 
The grammar scoring acts on the output of the \ASR, namely, $P_s$ and the gold truth $P_g$ (details mentioned later). This is done by displaying the paragraph $P_d$ generated by an LLM using prompt engineering.
%
We would like to emphasize that we 
do not focus on oral fluency and pronunciation (red dotted lines in Fig. \ref{fig:e2e-system}) which is common in \sla\ systems in this paper. Further, we do not delve into literature     
to focus on the proposed \sla\ system; an implementation is shown in Fig. \ref{fig:web_interface}.

\subsection{Generation of Paragraph}

A sample $P$ generated by prompting a LLM \cite{openai2023gpt35} is shown in Fig.  \ref{fig:sample-p}, \ref{fig:generated-p1} and \ref{fig:generated-p2}. The tags "<grammar> </grammar>" correspond to the words or phrases that are to be evaluated for grammar. 
The tag "<correct> </correct>" shows the correct choice. 
%
The correct choice of grammar usage is 
\textbf{studying} 
corresponding to \textbf{study/studied/studying} displayed to the student.
%
    In practice, both $P_d$  (Fig.  \ref{fig:sample-Pd}) and $P_g$ (Fig.  \ref{fig:sample-Pg}) can be obtained by a simple text parser applied on  $P$ (Fig.  \ref{fig:sample-p}).

\begin{figure}
\fbox{ \small
   \begin{subfigure}{0.5\textwidth}
    For <grammar><correct>a</correct>/an/the</grammar> student, <grammar>study/ studied/<correct>studying</correct></grammar> poetry can be a roller coaster ride. \myexchange{This journey <grammar><correct>is punctuated</correct>/ punctuates/ punctuated</grammar> by moments of profound appreciation <grammar>with/<correct>for</correct>/from</grammar> simpler pieces and intermittent frustration with more complex works. Some poems <grammar>were/ have been/<correct>are</correct></grammar> just plain confusing and no amount of re-reading  <grammar>seeming/<correct>seems</correct>/is seeming</grammar> to help decipher <grammar><correct>the</correct>/an /a</grammar> intended meaning. The puzzlement <grammar><correct>that</correct>/those/ these</grammar> results from such <grammar>institutions/<correct>instances</correct>/instigations</grammar>}{{\bf <snip>}} can be both vexing and <grammar><correct>demotivating</correct>/motivating/enthusing </grammar>.
    \caption{A paragraph generated by prompting a LLM ($P$).}
    \label{fig:sample-p}
    \end{subfigure}
    }
%
\fbox{ \small
\begin{subfigure}{0.5\textwidth}
    For \textbf{(a/an/the)} student, \textbf{(study/studied/studying)}  poetry can be a roller coaster ride. This journey \textbf{(is punctuated/punctuates/punctuated)} by moments of profound appreciation \textbf{(with/for/from)}simpler pieces and intermittent frustration with more complex works. Some poems \textbf{(were/have been/are)} just plain confusing and no amount of re-reading  \textbf{(seeming/seems/is seeming)} to help decipher \textbf{(the/an/a)} intended meaning. The puzzlement\textbf{ (that/those/these)} results from such \textbf{(institutions/instances/instigations)} can be both vexing and \textbf{(demotivating/motivating/enthusing)}.
    \caption{Paragraph displayed to the student ($P_d$).}
    \label{fig:sample-Pd}
    \end{subfigure}
    }
%
\fbox{ \small
\begin{subfigure}{0.5\textwidth}
    For \textbf{a} student,\textbf{ studying}  poetry can be a roller coaster ride. This journey \textbf{is }\textbf{punctuated} by moments of profound appreciation \textbf{for }simpler pieces and intermittent frustration with more complex works. Some poems \textbf{are} just plain confusing and no amount of re-reading  \textbf{seems} to help decipher \textbf{the} intended meaning. The puzzlement \textbf{that} results from such \textbf{instances} can be both vexing and \textbf{demotivating}.
    \caption{The grammatically correct paragraph ($P_g$).}
    \label{fig:sample-Pg}
     \end{subfigure}
     }
     \caption{A sample $P$ generated using an LLM along with $P_d$ used to display and $P_g$ used for grammar  assessment.}
     \label{fig:PPdPg}
\end{figure}



\subsection{Spoken Language Grammar Scoring}
The student is shown a paragraph $P_d$ on a web interface (Fig. \ref{fig:web_interface}) containing $
|P_d| 
$ words in language $L$. Of the $|P_d|$ words, a small subset of words
$G_w$ ($\in P_d, \ll P_d$) 
help determine the student’s grammar proficiency. 
%
The student ($s$) is given time to familiarize themselves with $P_d$ and then reads it into a microphone, generating the audio $S(t)$. 
The \sla\ system performs grammar scoring 
 in the following steps.
\begin{enumerate}
    \item[\#1] 
    Building a customized LM (\clm) specific to the paragraph $P$ to enhance the performance of the \ASR\  ($\text{\asrclm}$). 
    %
    Let 
    $P_s = \text{\asrclm}(S(t))$ be the transcript of $S(t)$.
\item[\#2] Compute the grammar score  ($S^{s}_{g}$) 
\begin{enumerate}
\item While maintaining the sequence of the words in $P_d$ and $P_s$, we create a set $ p_1 = \{ w \in P_d \mid w \notin P_s \} $ of words that are in $P_d$ but not in $P_s$.
\item Create 
$ p_2 = \{ w \in G_w \mid w \notin p_1 \}$.
\item The grammar score, $S^s_g= |p_2|$ is the cardinality of the set  $p_2$. Note that $p_2$ is a set of all the correctly spoken grammar words by the student.
\end{enumerate}
\end{enumerate}
In effect, the \sla\ of grammar 
takes $S(t)$, $P_d$, and $G_w$ as input and produces a score $S^s_g$. Namely,
\begin{equation}
    S^s_g = \text{\gs}(P_s, P_d, G_w)
    \label{eq:gs_model}
\end{equation}
where, $P_s = \text{\asrclm}(S(t))$.
    As an example, $G_w$ = \{a, studying, punctuated, for, are, seems, the, that, instances, demotivating\} for the paragraph shown in Fig.  \ref{fig:PPdPg} and $|P_d| = 61$.

\subsection{Speech to Text (\ASR)}
 The most crucial block is the \ASR, which converts the spoken paragraph $S(t)$ into text $P_s$ (see Fig. \ref{fig:e2e-system}) because \ASR\ \add{outputs} are erroneous \cite{10.5555/3061053.3061253} 
 leading to an error in grammar assessment. 
%
%
Let ${}^\error P_s$ be the true transcript (human transcribed) of 
$S(t)$.  Let $\e_s$ be the error due to \ASR, generally captured as the word error rate\cite{junqua1996robust} (\wer) between $P_s$ and ${}^\error P_s$, 
\begin{equation}
    \e_s = \text{\wer}(P_s, {}^\error P_s).
    \label{eq:asr_error}
\end{equation} 
%
Unless $\e_s = 0$, the audio grammar assessment score $S^s_g$ would be different from the true grammar assessment score, 
\begin{equation}
    {}^\error S^s_{g} = \text{\gs}(P_s^\error, P_d, G_w). 
    \label{eq:estimated_score}
\end{equation}
The error in grammar scoring due to an error ($\e_s$) in \ASR\  
is 
\begin{equation}
    \e_g = |S^s_g - {}^\error S^s_{g}|.
    \label{eq:grammar_error}
\end{equation}
We \myexchange{stress}{hypothesize} that 
in addition to the way \gs\ is determined (\ref{eq:gs_model}), 
the construction of \clm\,
tightly coupled with 
the assessment paragraph $P$ performs better than even the state-of-the-art \ASR\ (we use \sotaasr\cite{openai2022whisper} in our experiments). This is due to the fact that a LM plays a significant role in improving the accuracy of an \ASR\ engine. While 
\sotaasr\ 
is 
trained on extremely large and varied sets of text data, they are likely to {\em lack} 
grammatically incorrect sentences. 

As an illustration (see 
Fig.  \ref{fig:sen-var}) 
there are three possible options for both the preposition (a/an/the) and 
the verb (study/studied/studying). Hence, the total number of possible sentences using all options is nine. Most of these (eight of the nine) sentences will rarely occur, in any text databases since they are grammatically incorrect. 
Hence, text corpora used for training \sotaasr\ 
 will not include these sentences. Shallow fusion is the most popular approach to combine pre-trained \ASR\  model and LM \cite{e2e:survey}. Shallow fusion can be expressed mathematically as:
\begin{equation}
    \mathrm{score}(\C|\X) = \mathrm{log}\left (\P(\C|\X) \right)+\gamma\cdot \mathrm{log}(\P(\C))
\end{equation}
where $\C$ is the spoken paragraph, $\P(\C|\X)$ is acoustic score, $\gamma$ is a scaling factor and $\P(\C)$ is LM score. If $\C$ is not present in the training text, then $\P(\C) = 0$, which will make $\mathrm{score}(\C|\X)$ very small. This results in the \ASR\  choosing the {\em grammatically correct sentence instead of the spoken wrong sentence}. However, a \clm\ \cite{heafield2011kenlm} 
can, easily, be trained to include all possible variations (including the wrong ones) of the sentence to mitigate this. This is the reason for our belief that an \ASR\  with a custom-built LM (\asrclm) can be far more accurate than any state-of-the-art \ASR\  with a general-purpose LM.
\begin{figure}
\fbox
{
\begin{subfigure}{0.475\textwidth}
    \small For {(a/an/the)} student, {(study/studied/studying)}  poetry can be a roller coaster ride. 
    \caption{Sample sentence displayed to the student}
    \label{fig:sample-sen}
    \end{subfigure}
    }
\fbox
{
\begin{subfigure}{0.475\textwidth}
\small
    \begin{enumerate}
        \item For \textbf{a} student, \textbf{study} poetry can be a roller coaster ride.
        \item For \textbf{an} student, \textbf{study} poetry can be a roller coaster ride.
        \item For \textbf{the} student, \textbf{study} poetry can be a roller coaster ride.
        \item For\textbf{ a} student, \textbf{studied} poetry can be a roller coaster ride.
         \item For\textbf{ an} student, \textbf{studied} poetry can be a roller coaster ride.
          \item For\textbf{ the} student, \textbf{studied} poetry can be a roller coaster ride.
          \item {\em For \textbf{a} student, \textbf{studying} poetry can be a roller coaster ride.}
        \item For \textbf{an} student, \textbf{studying} poetry can be a roller coaster ride.
        \item For \textbf{the} student, \textbf{studying} poetry can be a roller coaster ride.
    \end{enumerate}
    \caption{Sentences (correct in italics) expected from the student.} 
    \label{fig:sample-var}
     \end{subfigure}
     }
     \caption{Sample sentence (a) and expected variations (b).} 
     \label{fig:sen-var}
\end{figure}

\section{Experimental Analysis}
\label{sec:experimental-results}

We first 
\add{describe how to generate a unique assessment paragraph}
$P$ 
\add{for each student} using \GenAI. This \add{ensures} that the students cannot be coached for the assessment. 
\add{Subsequently}, we experiment
with an instance of $P$ 
to validate the use of an \ASR\ engine equipped with a custom-built LM based on the generated paragraph, namely, \asrclm. 


\subsection{Generating $P$ using \GenAI}

We adopt 1-shot 
learning prompting style for generating new paragraphs ($P_1, P_2, \cdots$) as described in Fig.  \ref{fig:gen-paragraph}. 

\begin{figure}[!htb]
\fbox{\small \vbox{    
\noindent\textbf{\#1 User:} """ 
\myexchange{For <grammar><correct>a</correct>/an/the</grammar> student, <grammar>study/ studied/ <correct>studying</correct> </grammar> poetry can be a roller coaster ride. This journey <grammar><correct>is punctuated</correct>/ punctuates/ punctuated</grammar> by moments of profound appreciation <grammar>with/ <correct>for</correct>/ from</grammar> simpler pieces and intermittent frustration with more complex works. Some poems <grammar>were/ have been/ <correct>are</correct></grammar> just plain confusing and no amount of re-reading  <grammar>seeming/ <correct>seems</correct>/is seeming</grammar> to help decipher <grammar><correct>the</correct>/ an /a</grammar> intended meaning. The puzzlement <grammar><correct>that/ those/ these</grammar> results from such <grammar>institutions/<correct>instances</correct>/ instigations</grammar> can be both vexing and <grammar><correct>demotivating</correct>/ motivating/enthusing</grammar>.}{$P$}
"""  \{Sample $P$ in Fig.  \ref{fig:sample-p}.\}
\noindent \myexchange{The text enclosed between """ is a paragraph which is used for assessing vocabulary and grammar. 
Note that for grammar assessment there are 3 choices between the tags <grammar> and </grammar> and the correct answer is shown within the tags <correct> and </correct>.}{Generate paragraphs like $P$. One <correct></correct> tag within <grammar> </grammar> tags. Each <grammar> tag has three options separated by "/".
}


\noindent \textbf{\#1 \LLM:} \myexchange{Yes, that is clear.}{} Thank you for providing the specific format and instructions. The grammar choices are marked within <grammar>, with the correct option indicated using <correct>. 

\noindent\textbf{\#2 User:} \myexchange{I want you to generate a paragraph that can be used for grammar assessment. Make sure there are no more than 8 sentences in the generated paragraph. Only one attribute is to be tested in any given sentence. \myexchange{Have 4 vocabulary and 4 grammar assessments.}{}}{Generate a paragraph similar to the example shown.}




\noindent \textbf{\#2 \LLM}: 
%
{$P_1$} 
\{Generated paragraph \myexchange{for evaluation}{} (Fig.  \ref{fig:generated-p1})\}

\noindent \textbf{\#3 User}: Generate \myexchange{just the paragraph,}{} use subject "learning physics is easy".

\noindent \textbf{\#3 \LLM}: $P_2$  \{Generated paragraph shown in Fig.  \ref{fig:generated-p2}\}
}}
\caption{$1$-shot learning prompting to generate new $P$.}
    \label{fig:gen-paragraph}
\end{figure}
A wide variety of $P_n$'s can be generated using the prompt {\em "Generate just the paragraph. With subject <subject>."} 
%
\begin{figure}
\fbox{ \small
   \begin{subfigure}{0.45\textwidth}    
    {
{\begin{small}{In <grammar><correct>an</correct>/a/the</grammar> bustling city, <grammar>exploring/ explored/ <correct>exploration</correct> </grammar> can be an exciting adventure. \myexchange{This journey <grammar><correct>is marked</correct>/ marks/ marked</grammar> with moments of discovery <grammar><correct>amidst</correct>/ among/ between </grammar> the chaos and occasional frustration with crowded streets. Some neighborhoods <grammar>were/ <correct>have been</correct>/ are</grammar> vibrant and diverse, and no amount of time <grammar>seems/ <correct>seem</correct>/seeming</grammar> to fully explore <grammar>all/ <correct>the</correct>/ any grammar> hidden gems.}{{\bf <snip>}} The challenge <grammar><correct>that</correct>/ those/these</grammar> comes from such <grammar>adventures/<correct>explorations</correct>/explorers </grammar> can be both thrilling and <grammar>eye-opening/<correct>exhausting</correct>/ insightful</grammar>.}\end{small}}
}
    \caption{A paragraph generated by prompting  \LLM\ ($P_1$).}
    \label{fig:generated-p1}
    \end{subfigure}
    }
    \fbox{\small
   \begin{subfigure}{0.45\textwidth}    
    {
   For <grammar><correct>an</correct>/a/the</grammar> physics enthusiast, <grammar>studying/ studied/ <correct>studying</correct> </grammar> physics can be a fascinating journey. \myexchange{This educational experience <grammar><correct>is filled</correct>/fills/filled</grammar> with moments of enlightenment <grammar>from/ <correct>by</correct>/ with</grammar> fundamental principles and occasional frustration with complex equations. Some concepts <grammar>were/ have been/ <correct>are</correct></grammar> inherently clear, and no amount of practice <grammar><correct>seems</correct>/seem/seeming</grammar> to dull <grammar><correct>the</correct>/ an /a</grammar> curiosity.}{{\bf <snip>}} The understanding <grammar><correct>that</correct>/ those/ these</grammar> comes from such <grammar>endeavors/<correct>pursuits</correct>/ explorations</grammar> can be both empowering and <grammar><correct>rewarding</correct>/ challenging/exciting</grammar>.
    }
    \caption{A paragraph generated by prompting \LLM\ ($P_2$).}
    \label{fig:generated-p2}
    \end{subfigure}
    }
    \caption{Paragraph's generated by prompting \LLM.}
    \label{fig:generated-ps}
    \end{figure}
This allows for the generation of a completely new paragraph in the desired format; the sample generated $P$ shown in 
Fig.  \ref{fig:generated-p1}, and \ref{fig:generated-p2}. 

\subsection{\ASR\ performance}
We used \sotaasr\ speech recognition engine 
and a Kaldi-based \ASR\  with a custom-built LM (\asrclm) for comparison. The acoustic model of the Kaldi \ASR\  was trained on 960 hours of speech data from Librispeech database \cite{pana:librispeech}. The custom LM was trained on the text comprising all possible variations of the given sentences (example Fig. \ref{fig:sample-var}).
%
We recorded speech corresponding to all variations of the below sentence,
"It (\textbf{was}/is/am) a late afternoon probably (\textbf{on}/in/of) the 15th of February, 2019.  (I and my friend/\textbf{my friend and I}) (was/\textbf{were}/will be) walking on the footpath (\textbf{in}/inside/into) central Bangalore.\myexchange{My friend is (a/an/the) RBI employee who is also (a/an/the) amateur photographer.  Although he has other academic interests he by profession is (a/an) IT engineer.  We walked (into/in) the Cubbon park. My friend is quick in everything he (do/does). He (speak/speaks) fast, (walk/walks) fast, (eat/eats) fast and (is/was) always in a hurry. As we (begin/began/begun) walking inside, he as usual with his hurried gait (tripped/trips) on a stone and fell forward.}{}"
  namely, $3 (\text{was/is/am}) \times 3 (\text{on/in/of}) \times 2 (\text{I and my friend/my friend and I}) \times 3 (\text{was/were/will be}) \times 3 (\text{in/inside/into}) = 162$ utterances. We found that \asrclm\ 
was able to exactly transcribe the utterance (even when there was an error in grammar) while 
\sotaasr\ "corrected" the grammatical error.
%
%
Table \ref{tab:sample-ctlm} shows two examples where \asrclm\ 
accurately recognizes the spoken words, regardless of grammatical correctness, while \sotaasr\ 
falls short. In the first example (Table \ref{tab:sample-ctlm}) the article {"a"} was replaced by {"the"} by  \sotaasr\ while in example two, the article {"a"} was not recognized by \sotaasr. 
Overall, the ability of \text{\asrclm} to recognize what was spoken is \myexchange{$73.88\%$}{$84.7\%$} while that of \sotaasr\ 
was \myexchange{$40.12\%$}{$46\%$}. The performance was computed on  $137$ utterances; $25$ of the $162$ utterances were discarded because of noise. The poor accuracy of the SOTA \ASR\ highlights the need for a 
\clm-\ASR\ for the purpose of \sla\ of grammar. 
%
%
 %
 	\begin{table}[!htb]
 	    \centering
 	    \begin{tabular}{|c|p{6cm}|} \hline
 	 $S(t)$     &  It was a late afternoon probably on the 15th of February  2019  \mfai\ were walking on the footpath in central Bangalore\\
 	       \sotaasr  & It was {\color{red}{the}} late afternoon probably on the 15th of February 2019  \mfai\ were walking on the footpath in central Bangalore \\
            $\text{\asrclm}$      &"It was a late afternoon probably on the 15th of February 2019 \mfai\ were walking on the footpath in central Bangalore". \\ \hline\hline
          $S(t)$     &    It am a late afternoon probably on the 15th of February 2019 \mfai\ was walking on the footpath into central Bangalore \\	  
           \sotaasr  &
           It am {\color{red}{\sout{a}}} early after noon probably on 15th February 2019  \mfai\ was walking on the footpath in central Bangalore 		\\ 
          $\text{\asrclm}$ & It am a late afternoon probably on the 15th of February  2019  \mfai\ was walking on the footpath into central Bangalore \\ \hline
 	    \end{tabular}
 	    \caption{Sample $S(t)$. 
      \ASR\ errors, marked in red.}
 	    \label{tab:sample-ctlm}
 	\end{table}

\add{To the best of our knowledge, a standard speech dataset for spoken grammar assessment with manual annotations of grammatical errors in conversational or read speech is currently unavailable. To evaluate our SLA system, we used an in-house dataset consisting of audio recordings from $17$ students speaking a generated paragraph, which was manually assessed by a linguist} to mark the grammar score (${}^\error S^s_{g}$). We used both \sotaasr\ and \asrclm\ to convert the spoken paragraph to text 
and compute $S^s_g$. 
The error in assessment is captured in parenthesis for each student in Table \ref{tab:compare}. Larger grammar assessment errors ($\e_g = 20$) due to \sotaasr\ are observed compared to $\e_g = 3$ for a custom-built LM \ASR\ (\asrclm). 
 %
\begin{table}[!htb]
    \centering
    \begin{tabular}{|c|c|c|c|} \hline
    \multirow{3}{*}{Student}& \multicolumn{3}{c|}{Grammar Assessment} \\ \cline{2-4}
    & \sotaasr\ &  \asrclm\  & \multirow{2}{*}{${}^\error S^s_{g}$}\\ \cline{2-3}
     & $S^s_g (\e_g )$  &  $S^s_g (\e_g )$  & \\ \hline
       \Della  &  14 (1) & 15 (0)& 15 \\
        \Upasana & 11 (1) & 11 (1) & 10 \\
        \Samina & 11 (2) & 9 (0)& 9 \\
        \Ashwini & 12 (1) & 13 (0)& 13 \\
        \sSunilzerozero & 12 (1) & 12 (1) & 13\\
        \sSunilzeroone & 10 (2) & 12 (0) & 12 \\
        \sSunilzerotwo & 6 (2) & 8 (0) & 8 \\
\onesevenonefivefivesevenzeroone & 15 (3) & 12 (0) & 12 \\ 
\fiveonenineseversixsixzeroone & 15 (1) & 16 (0) & 16\\
\fiveonetwosevensixsixzerotwo & 3 (0) & 3 (0) & 3\\
\fiveonenineseversixsixzerothree & 6 (2) & 8 (0) & 8\\ 
\fiveonenineseversixsixzerofour & 10 (2) & 12 (0) & 12\\ 
\eightfiveeightseventhreezerozeroone &15 (1) & 15 (1)& 16\\
\prajwalzeroone & 14 (1) & 15 (0) & 15\\ 
\rajulzeroone & 14 (0) & 14 (0) & 14\\
\Thambidurai & 13 (0) & 13 (0) & 13\\
\hline
Total & (20) & (3) & - \\ \hline
    \end{tabular}
    \caption{Use of \sotaasr\ and \asrclm\ for grammar assessment. $\e_g$  computed using (\ref{eq:grammar_error}).}
    \label{tab:compare}
\end{table}




\section{Conclusions}
\label{sec:conclusions}

Language proficiency assessment is a common requirement for \ltwo\ speakers of English. 
There exist several \sla\ tools to assess pronunciation and oral fluency but none of them venture into assessing language grammar, instead, they depend on \wla\ systems.
We designed and implemented a practical, scalable and robust \sla\ system to assess grammar. The design, to display the paragraph with options, made sure the audio obtained for assessment had no spontaneous speech characteristics like filler words, or repetitions and resembled "read" speech thereby enhancing the \ASR\ performance. Additionally, the use of a custom LM in \asrclm\ leads to improved \ASR\ performance, resulting in robustness in grammar assessment.  The use of LLM enables the generation of paragraphs that are largely non-repetitive thereby making the proposed system hard to be memorized by students. 
We can observe that the grammar scoring mechanism, 
by design, is not affected by \ASR\ mis-recognition of 
non $G_w$ words. 



\bibliographystyle{ieeetr}
\bibliography{mybib,eusipco_oral-fluency}


\end{document}